%% file: magazine_ArXiv_R2.tex
\documentclass[journal]{IEEEtran}
\pdfoutput=1
\usepackage{cite}
\usepackage{amsmath,amssymb,amsfonts}
\usepackage{graphicx}
\usepackage{textcomp}
\usepackage{xcolor}
\usepackage{algorithmic}
\usepackage{algorithm}
\usepackage{array}
\usepackage[caption=false,font=normalsize,labelfont=sf,textfont=sf]{subfig}
\usepackage{stfloats}
\usepackage{url}
\usepackage{verbatim}
\usepackage{cite}
\usepackage{booktabs}
\usepackage{tikz}
\usetikzlibrary{arrows,automata,matrix,fit,positioning,calc}
\usepackage{balance}
\usepackage[normalem]{ulem}

\begin{document}


\def\ArXivVersionCopyright{0}   

\def\IEEEVersion{0}     

\def\KillMyDarlings{0}    

\def\mycmdextendedpaper{0} 
\title{Design Principles for Model Generalization and Scalable AI Integration~in~Radio~Access~Networks}

\author{Pablo Soldati, Euhanna Ghadimi, Burak Demirel, Yu Wang, Raimundas Gaigalas, Mathias Sintorn\\ \IEEEmembership{Ericsson AB, Stockholm, Sweden,
email:\{name.surname\}@ericsson.com}
\if\ArXivVersionCopyright1
\thanks{\copyright~20xx IEEE. Personal use of this material is permitted. Permission from IEEE must be obtained for all other uses, in any current or future 
media, including reprinting/republishing this material for advertising or promotional purposes, creating new collective works, for resale or redistribution to servers or lists, or reuse of any copyrighted component of this work in other works.}
\fi}



\maketitle

\begin{abstract}
Artificial intelligence (AI) has emerged as a powerful tool for addressing complex and dynamic tasks in radio communication systems. Research in this area, however, focused on AI solutions for specific, limited conditions, hindering models from learning and adapting to generic situations, such as those met across radio communication systems. 

This paper emphasizes the pivotal role of achieving model generalization in enhancing performance and enabling scalable AI integration within radio communications. We outline design principles for model generalization in three key domains: environment for robustness, intents for adaptability to system objectives, and control tasks for reducing AI-driven control loops. Implementing these principles can decrease the number of models deployed and increase adaptability in diverse radio communication environments.
To address the challenges of model generalization in communication systems, we propose a learning architecture that leverages centralization of training and data management functionalities, combined with distributed data generation. We illustrate these concepts by designing a generalized link adaptation algorithm, demonstrating the benefits of our proposed approach.
\end{abstract}


\input{components_arXiv_R2/Section1.tex} 

\input{components_arXiv_R2/Section2.tex} 

\input{components_arXiv_R2/Section3.tex} 

\input{components_arXiv_R2/Section4.tex} 

\input{components_arXiv_R2/Section5.tex} 

\input{components_arXiv_R2/Section6.tex} 

\input{components_arXiv_R2/Section7.tex} 

\input{components_arXiv_R2/Section8.tex} 

\bibliographystyle{IEEEtran}

\if\IEEEVersion1
\bibliography{references_very_short.bib}
\input{components_arXiv_R2/Biographies.tex} 
\else
\bibliography{references_short.bib}
\fi

\balance
\vfill
\end{document}

%% file: components_arXiv_R2/Section1.tex
\section{Introduction}

Achieving model generalization is a fundamental design paradigm of Machine Learning (ML) and Artificial Intelligence (AI). Model generalization refers to the ability of a trained model~to~perform well on new inputs unseen during training, making AI applications reliable to cope with the diversity of real-world scenarios \cite{DeepLearning:15}.

For prediction problems, techniques like regularization, data augmentation, and transfer learning, helped traditional ML avoiding overfitting and extract more robust, generalizable patterns \cite{GenDeepLearn:2022}. However, only the proliferation of high-performance computing resources enabled fully capitalizing on deep learning algorithmic breakthroughs and adopting larger and more advanced models to enhance generalization in complex prediction tasks by learning intricate patterns from vast volumes of data. In natural language processing, transformers \cite{Vaswani:2017} and graph attention networks \cite{GAT:2018}  have demonstrated superior performance compared to traditional recurrent networks. This underscores the critical role of attention layers in improving model generalization. These techniques laid the groundwork for Generative Pre-trained (GPT) models and Large Language Models (LLMs) that drive generative AI today.

Addressing complex control problems with AI, involving decision-making and adaptability in dynamic environments, presents additional challenges to model generalization. In this context, model generalization can be rephrased as the ability of an agent to effectively apply the learned knowledge and policies to novel environments or tasks unseen during training; see, e.g., \cite{FMB:18, RLGen:19, reed2022a, KZG+:23}.  With reinforcement learning (RL) algorithms, for instance, generalization is often studied in the context of Atari games, with an agent trained on a subset of games, each with distinct dynamics and objectives, and tested on unseen games. Like prediction tasks, achieving model generalization for control typically requires a large dataset generated by the agent interaction with simulated or real environments.  

However, these breakthroughs in AI model generalization would only be possible with a scalable architecture capable of efficiently handling the data volumes required to train large models and parallelizing the computations required by the training process across many distributed resources. For instance, the latest LLM from Meta, LLAMA2, comprises 7 to 70 billion parameters trained over two trillion data tokens using thousands of powerful GPUs.

Prediction and control are critical elements of many functionalities of radio communication systems for fast real-time operations, like channel estimation and signal demodulation at the physical layer (L1), Radio Resource Management (RRM) at the medium access layer (L2), load prediction and mobility decisions at the network layer (L3), as well as slower network optimization and planning operations, like cells shaping or re-configuration. The ability of AI/ML to excel in prediction and control tasks sparked a surge of research aiming at replacing the traditional rule-based designs of virtually any network functionalities with models or control policies learned from data. However, model generalization and scalability of learning have largely been neglected in this context.

This paper argues that AI model generalization is the cornerstone for a scalable integration of AI in radio communication networks. We identify design principles for model generalization in three domains relevant to communication systems: environment, intents, and control. We discuss the challenges that achieving model generalization poses to the scalability of training and data generation, and we propose an architecture that addresses these challenges effectively and suits radio communication systems well. Hereafter, we frame the discussion in the context of Radio Access Networks (RAN); however, concepts presented in this paper apply to other radio communication systems. 

%% file: components_arXiv_R2/Section2.tex
\section{Why model generalization?}\label{sec:motivation}
Current implementations of AI in radio communication systems, such as RAN, predominantly use artificial narrow intelligence or “narrow AI,” characterized by techniques specifically designed and trained for a defined task. For instance, AI solutions meticulously designed and trained to address individual network tasks in L1 (e.g., encoding and decoding information), RRM tasks in L2 (e.g., power control, scheduling, link adaptation (LA), etc.), mobility optimization, network configuration in L3, etc. 

Task specialization of AI proved successful in many disciplines, and may facilitate replacing some network functionalities with data-driven design. For instance, GPT models combined with discriminative fine-tuning to specific tasks outperformed state-of-the-art in many NLP applications, such as natural language inference, question answering, semantic similarity, text classification, etc.,~\cite{GPT1:18}. However, even within the context of task-oriented AI, model generalization remains a design cornerstone. 

AI research in communication systems, on the other hand, often overlooks the crucial role of model generalization, striving instead for models rigorously designed, trained, and evaluated under specific network conditions, deployments, or radio environment. This form of model specialization can pose significant challenges to AI integration in real-world RAN systems, such as: 

\begin{itemize}
    \item \textbf{Model proliferation}: Training numerous AI models for a specific RAN function (e.g., link adaptation), each tailored to localized network environments (e.g., a particular radio cell), incurs substantial costs and proves impractical for scalability and lifecycle management. 
    \item \textbf{Limited Robustness}: Specialized models face challenges adapting to changes in network deployment and radio environments, as they are biased towards the limited conditions seen during training. This necessitates regular model updates throughout the network, impacting overall performance consistency. 
    \item \textbf{Design fragmentation}: RAN functionalities exhibit strong interdependencies, especially within the same protocol layer (e.g., link adaptation and scheduling). A fragmented design with multiple coexisting AI solutions, each addressing an individual RAN functionality, operating independently and concurrently at the same timescale, and learning from the same environment (e.g., the same cells), can introduce instability to the learning process.
\end{itemize}

To the contrary, achieving model generalization can overcome these problems, allowing a single model or control policy deployed to replace a RAN functionality to consistently operate across the network, a key factor for scalable and efficient AI integration in RANs. It is crucial to clarify, however, that with model generalization we do not suggest a generalist AI~\cite{reed2022a} that handles all tasks. Rather, it involves developing adaptable, task-oriented AI models capable of effectively operating in diverse conditions.

The following sections delve into intricacies of attaining model generalization in RAN, starting with an essential question: \emph{What does model generalization entail in the context of RAN?}

%% file: components_arXiv_R2/Section3.tex
\begin{figure}[!t]
\centering
\includegraphics[width=1\columnwidth]{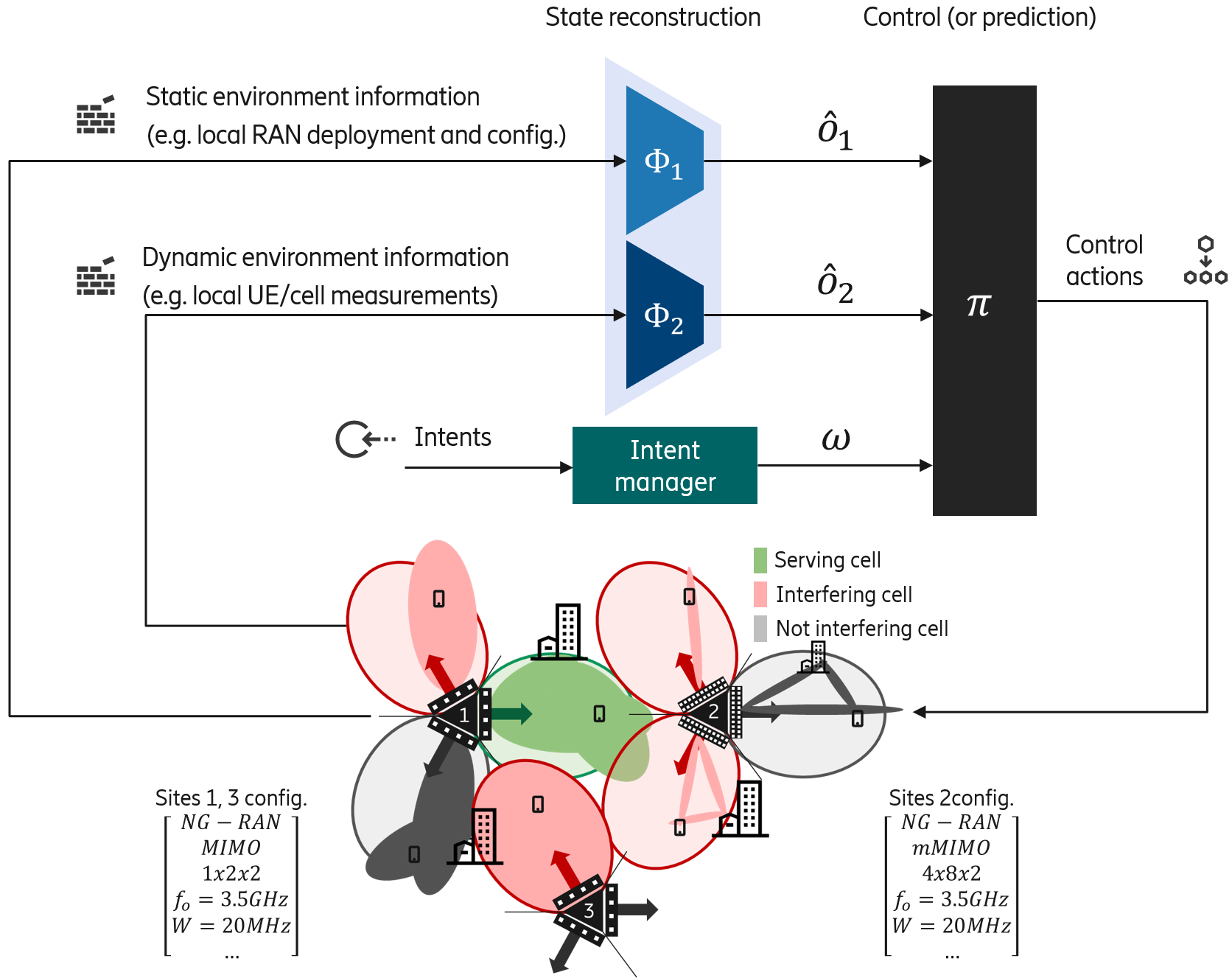}
\caption{A framework for model generalization in radio systems over network environment, intents, and control.}
\label{fig:1}
\end{figure}

\section{Design principles for model generalization}\label{sec3:model_gen}

To address this question, we identify three primary design domains pertinent to model generalization in communication systems: \emph{RAN environment} to enable ubiquitous deployment of a single model with consistent performance; \emph{RAN intents} to handle and optimize different key performance indicators (KPIs) with a single model at runtime; and \emph{RAN control} to reduce the number of co-existing AI-driven control loops acting on dependent RAN functionalities or parameters. 

We frame our discourse primarily on RAN control applications, although insights provided for model generalization over RAN environment also apply to prediction scenarios. 

\subsection{Model generalization over environment}\label{subsec3a:gen_over_env}

Paraphrasing \cite{DeepLearning:15}, generalization over the RAN environment implies the model's ability to adapt to new network conditions, deployments, radio environments,~etc.,~unseen~in~training.
Attaining this form of generalization for control or prediction requires feature robustness and robustness to uncertainty~\cite{RLGen:19}. 

Feature robustness refers to model's resilience to inputs aleatory uncertainty. This property requires an effective input features design and is crucial to generalize in environments with high-dimensional state-space like RAN systems. 
In contrast, robustness to uncertainty assesses a model's capacity to handle epistemic uncertainties from various sources, e.g., noisy data, imprecise measurements, and environmental changes, without substantial performance degradation. Epistemic uncertainty renders many RAN control problems unobserved Markov Decision Processes, where the current state lacks the essential information for making optimal decisions within the prevailing MDP. Achieving model robustness in RAN involves two key aspects: precise state reconstruction to accurately reflect the environment's dynamics despite uncertainties; and training a model across diverse environments.  Here, we discuss the former and address the latter in Sec.~\ref{sec4:impact_training}.  

Unlike AI in fully observable environments (e.g., Atari, Go, NLP), state reconstruction in RAN faces challenges due to partial observability and heterogeneous aging of measurements. Partial observability is a limitation common in sensorial systems. RAN entities, like access nodes and radio cells, perceive the state-space regulating a RAN functionality only within a local surrounding, either through direct measurements or with information from user devices or neighboring nodes, as Figure~\ref{fig:1} illustrated for the green cell. A distinctive aspect of RAN systems, however, is the uncertain availability of neighbors' measurements and information. For instance, while 3GPP specifications provides procedures to request neighboring information (e.g., cell load), it may be temporarily unavailable or unsupported by some vendors. Often overlooked in simulation-based studies, accounting this unreliability is important for real-world AI applications.

Furthermore, state features of a RAN functionality may include diverse timescales, spanning static information (e.g., RAN deployment, cell configuration), to slow information (e.g., load or traffic variations) and fast dynamics (e.g., millisecond-level channel measurements). Consequently, the RAN functionality state reconstructed at a time $t$ is often an imprecise representation of the true state, with features providing historical snapshots affected by various aging factors.

To improve model generalization over RAN environment via feature robustness, we propose a state reconstruction design that separates static and dynamic environment information and applies different transformations, cf. $\Phi_1$ and $\Phi_2$, respectively, as shows in Figure~\ref{fig:1}. 

\begin{itemize}
    \item \textbf{Static environment information} shall capture the RAN deployment heterogeneity and its configuration, so a model can seemingly operate across the network. This encompasses network topology details, including deployment area type, location, orientation, and relationships of network sites or radio cells. It also contains specifics outlining their transmission technology and configuration (e.g., antenna array characteristics, carrier frequency, bandwidth, transmit power). Including user device characteristics, e.g. chipset type, antenna array, etc., allows a model to operate with heterogeneous user devices from different manufacturers. Although static or semi-static, this information substantially influences the radio environment, signals and observed state-space.
    \item \textbf{Dynamic environment information} pertains to the measurable state dynamics characterizing the specific RAN functionality that AI replaces. Sec.~\ref{sec:Evaluations}, for instance, provides an example for link adaptation which includes user path loss, channel quality indicators, hybrid automatic repeat request (HARQ) feedback, etc.
\end{itemize}

Applying different transformations $\Phi_1$ and $\Phi_2$ to static and dynamic environment information, enables us to consider different approaches to state reconstruction:

\subsubsection*{1. Domain expert approach} When $\Phi_1$ and $\Phi_2$ represent identity functions, the approach shown in Figure~\ref{fig:1} is equivalent to traditional feature engineering. This requires domain expert knowledge to carefully select static and dynamic information to reconstruct the observable state relevant to improve model generalization.

\subsubsection*{2. ML-based reconstruction}  An alternative approach involves learning an embedding characterizing the static environment information for potential reuse in various RAN functionalities. Suitable models $\Phi_1$ for this purpose include Graph Neural Networks (GNNs) \cite{Scarselli:2008}, and attention networks \cite{GAT:2018}. Graph representations effectively capture topological and configurational relationships among network entities. Meanwhile, attention layers can weigh different information types and their interrelations, eliminating extensive input pre-selection. These models can integrate with different designs of $\Phi_2$ for dynamic environment information, such as an identity function for traditional feature engineering or more advanced ML techniques like an autoencoder, which are typically used for dimensionality reduction or feature learning~\cite{DeepLearning:15,DeepLearningBook:16}.

\subsection{Model generalization over RAN control parameters}\label{subsec3b:gen_over_cont}
We define model generalization over RAN control as the model's ability to effectively handle multiple RAN control parameters concurrently, particularly when interdependent. This form of generalization offers two primary advantages: (1) minimizing coexisting AI-driven control loops and (2) potentially outperforming rule-based designs by replacing split decisions with end-to-end learned joint decisions. Although this approach may help reduce design fragmentation (cf. Sec.~\ref{sec:motivation}), it brings new challenges, notably the increased complexity in computing an acceptable control policy $\pi$ (cf. Figure~\ref{fig:1}) due to larger action-spaces and execute it within stringent real-time demands.

Traditional AI generalization for control aims at training a single agent to handle a set of tasks, each having independent dynamics and objectives. For example, recent RL studies use Atari games as distinct control tasks~\cite{reed2022a}. In this context, certain actions (like moving right/left/up/down) may be shared among some games, while others are game-specific. By employing methods like action masking~\cite{Nair:2015} and establishing a well-defined reward function, a single agent (i.e., one model) can be trained to \emph{individually play} various games, often surpassing human-level performance. 

Applying this approach to radio networks brings new challenges for both training and execution, in addition to those arising from dealing with larger action-spaces. 
A first challenge is the need to execute the  model multiple times, one per RAN functionality separately, masking parameters subsets for different tasks. For instance, if a generalist agent~\cite{reed2022a} were designed and trained to handle control power control and link adaptation parameters in L2, it would require to be executed twice, with each execution returning a subset of the transmission parameters related to a RAN functionality. This is computationally intensive within real-time requirements of communication systems (e.g., sub-millisecond timescale for L2 functionalities). 


A second challenge related to designing and training a generalist AI for RAN control, on the other hand, is the tight interdependencies among control parameters of different RAN tasks, where selecting control parameters for one task influence the optimization of control parameters of subsequent tasks. Parameters (i.e., actions) interdependence across tasks hinders learning an admissible control policy $\pi$ (cf. Figure~\ref{fig:1}) from individual tasks in parallel, as in~\cite{reed2022a}. These challenges are typically absent in problems studied in generalization literature, like Atari games.

\textbf{\textit{Example:}} Link adaptation, power control, beamforming, and resource scheduling are examples of tightly interconnected L2 functionalities. For instance, the scheduling time-frequency resources for a user transmission instance impacts the optimization of link adaptation parameters (i.e., modulation order, transport block size and rank) which depend on the channel quality. However, the resource allocation itself is affected by link adaptation parameters, as the choice of modulation order affects transmission robustness and the amount of information bits carried per resource element, consequently affecting the number of time-frequency resources necessary to fulfil a user transmission (e.g., in terms of information bits). RAN systems typically deal with these intricate relationships through intertwined rule-based algorithms and design simplifications. 

Contrary to advocating for a generalist AI~\cite{reed2022a}, we consider a pragmatic approach to address these challenges and avoid introducing computational complexity for diminishing returns. Firstly, we propose to categorize RAN control functions based on their timescales relying on the layered RAN protocol stack, with an AI agent handling multiple RAN operations within a protocol layer. This ensures that the model would handle control parameters operating at the same timescale. 

An approach to address dependencies across control parameters of multiple RAN functionalities is to optimize them simultaneously, i.e., treating them as a single task. However, a holistic approach involving the joint optimization of many RAN functionalities may be infeasible and unnecessary.

We advocate instead that \emph{not every RAN functionality benefits from AI}. Therefore, rather than applying AI to all (or many) RAN functionalities within a RAN protocol layer, we propose (1) to exploit expert domain knowledge to identify which RAN functionalities may most benefit AI, and (2) to apply a \emph{joint} data-driven design to such functionalities. 
Carefully selecting a subset of key RAN functionalities makes a joint AI design manageable, addressing tasks interdependence while also reducing complexity of larger action-space. It also enables to reduce design fragmentation (cf. Sec.~\ref{sec:motivation}), avoiding instability in the learning process that may occur if multiple coexisting AI solutions, each addressing an individual RAN functionality, operating independently and concurrently at the same timescale, tried to interact and learn from the same environment (e.g., within the same cells).

\textbf{\textit{Example:}} Certain L2 RAN functionalities, like power control, have near-optimal, low-complexity rule-based solutions, making AI unnecessary. In contrast, L2 functions like scheduling and link adaptation, which often rely on simplified designs to meet product constraints, may attain better performance with a joint AI design.

\subsection{Model generalization over intents}\label{subsec3c:gen_over_intents}

Intent-based operation is essential for building autonomous networks. The term “intent” refers to objectives, requirements, and constraints for a service or network operation, see, e.g., in 3GPP TS 28.312. However, different intents can lead into conflicts as they touch upon multiple, often competing, network KPIs. This calls for trade-offs among KPIs, resulting in a Pareto frontier wherein optimizing one KPI could compromise another. For instance, enhancing spectral efficiency might reduce transmission reliability and vice versa.

We define generalization over intents as the ability to learn a single model that balances competing objectives parameterized by user-defined preferences. AI research in communication predominantly focuses on optimizing individual intents, requiring separate models for different KPIs, thereby missing the advantages of holistic optimization across these trade-offs.
Figure~\ref{fig:1} showcases a framework proposing the training of one model (for each RAN function) that captures the Pareto frontier of multiple RAN intents. We realize this by coupling an intent manager with a Multi-objective RL (MORL) design, such as envelope Q-learning \cite{YSN:19}. The intent manager maps intents to preference values $\omega$, which, in turn, determine the significance of each reward component for the downstream MORL controller $\pi$. A single policy optimizing all reward components emerges by training this controller with varying preference values. The resultant model, $\pi$, can adjust in real-time to offer optimal KPIs tailored to specific scenarios, like diverse quality-of-service (QoS) demands (cf. Sec.~\ref{subsec:results_intents}).

%% file: components_arXiv_R2/Section4.tex
\section{Scalability via distributed learning}\label{sec4:impact_training}

\subsection{Training and data generation requirements}

Learning from various environments enhances robustness which, in turn, improves model generalization, as highlighted in\cite{RLGen:19}. Domain randomization achieves this by introducing \emph{environment randomness} during training to bridge the simulation-to-reality gap. A natural application of domain  randomization is improving AI model resilience against RAN environment uncertainties, e.g., training a model concurrently over parallel simulations with randomized parameters such as network deployment and configuration, traffic conditions, user device models, etc.

We extend this concept to generalization over RAN intents and control. Generalization over intents, which deals with model adaptability to various objectives, can be realized by simultaneously evaluating a multi-dimensional value function across a range of intents (Sec.~\ref{sec:Evaluations} describes an example). For control tasks, however, the generalization process involves learning across combined state-action spaces corresponding to joint RAN functionalities.

Generalization across these domains requires diverse training data, each tailored to specific domain requirements. Managing the diversity needed for generalization involves conducting extensive simulations and handling massive datasets, demanding a delicate balance between data generation and the computational capacity required for learning from incoming data. Next, we review distributed RL algorithms that demonstrated scalability in learning, while in Sec.~\ref{sec:scalable_learning} we discuss their application to RAN systems.

\subsection{Scalability in RL architectures}
Since the publication of the seminal DQN algorithm \cite{Mnih:2013}, several architectures progressively addressed the scalability and effectiveness of the learning in RL algorithms \cite{Nair:2015, Mnih:2016, Espeholt:2018, Horgan:2018, Kapturowski:2018}. Impala \cite{Espeholt:2018} achieved a first milestone combining off-policy learning with actor-learner decoupling into a distributed architecture that capitalizes on, rather than suffer from, the lag between when the actors generate actions and when the learner estimates the gradient. An architecture advocating actor-learner decoupling for RAN applications was independently proposed in \cite{Calabrese:2018}. Recent RL architectures, like Ape-X \cite{Horgan:2018} and R2D2 \cite{Kapturowski:2018}, have expanded on these principles, pushing the limits of learning scalability and effectiveness. 

These architectures exploit a key design principle: a \emph{single learner} supporting multiple \emph{distributed actors}, all contributing to learning a single RL policy from experience generated by many simulations in parallel. Therefore, learning is distributed as multiple actors independently generate experience from different environments in parallel, while training and data management and storage is centralized.
Asynchronous policy updates enable actors to evaluate policies across multiple environments simultaneously (e.g., in parallel simulations), improving data generation throughput and resulting in faster and more efficient learning. The R2D2 architecture \cite{Kapturowski:2018} outperformed traditional DQN, achieving $5-20$x performance boost and $2.5-50$x shorter training time.

%% file: components_arXiv_R2/Section5.tex
\section{Scalable integration of AI in radio networks}\label{sec:scalable_learning}

\begin{figure*}[!t]
\centering
\includegraphics[width=2 \columnwidth]{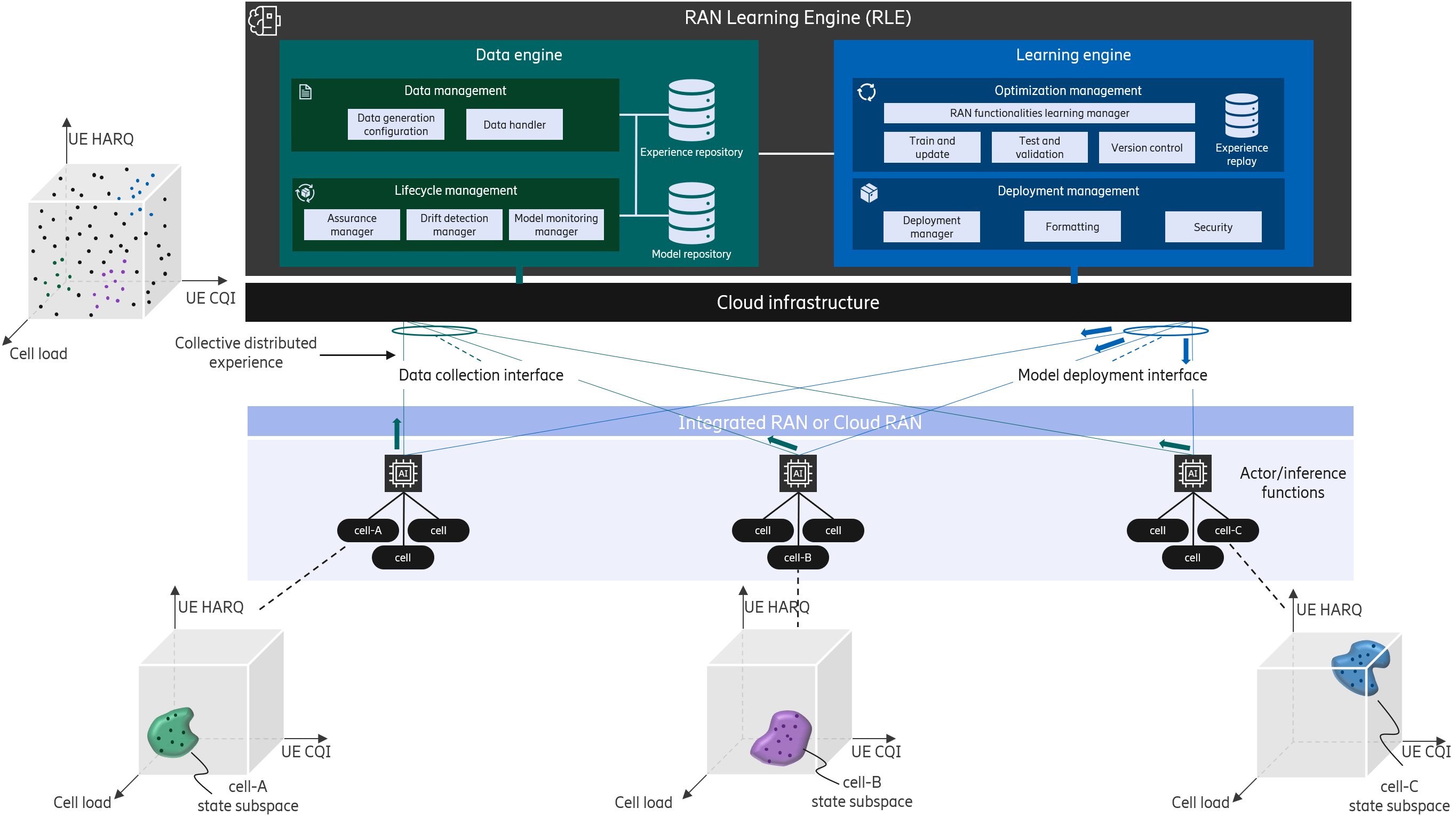}
\caption{An architecture for scalable integration of AI in RAN systems featuring centralized learning and data services to AI-driven network functionalities distributed in the underlying RAN.}
\label{fig:2}
\end{figure*}

Figure~\ref{fig:2} illustrates the functional elements of an architecture designed for scalable and distributed learning in RAN systems. This architecture draws inspiration from principles of distributed reinforcement learning developed for simulation environments. It features a centralized RAN Learning Engine (RLE), which provides learning services to AI-driven RAN functionalities hosted within distributed RAN entities, such as gNB-Central Unit (gNB-CU), gNB-Distributed Unit (gNB-DU), x/rApps, etc., responsible for control (acting) or prediction (inference) tasks.

Learning is distributed: each RAN entity samples data from its local environment and contributes to a collective knowledge pool, as shown in Figure~\ref{fig:2}. This boosts AI-driven system's robustness to uncertainties commonly encountered in real-world deployments. However, training and data management are centralized, enabling more efficient data collection and usage in large-scale RAN systems and it is vital to scalable learning from large collective knowledge pools.

For clarity, we divide the functionalities of the RLE into two main components: a learning engine and a data engine. This division simplifies the architecture and enables an efficient distribution of computational resources, ensuring the right balance between power and cost (cf. Sec.~\ref{subsec:resource_dimensioning}).

\subsection{The learning engine}\label{sec5:learning_engine}
The learning engine provides centralized AI learning services to AI-driven RAN functions requiring actuation or inference deployed in RAN entities. A supervisory management function oversees the instantiation and execution of learning and deployment pipelines for each RAN function. 

A learning pipeline includes services for training, updating, validating, and testing AI models and algorithm-specific components, such as experience replay memories for specific RL algorithms. Hence, the learning engine must host the necessary AI libraries and provide fast access to hardware accelerators. The learning pipeline further offers model version control during training and lifecycle management (LCM) operations, closely interacting with the data engine's model repository for storage and maintenance. The deployment pipeline delivers AI models to RAN entities hosting actor/inference functions. This may occur in two ways. For training RL applications, actors in the underlying RAN system may asynchronously pull AI models from the deployment manager for distributed policy evaluation. Alternatively, during LCM operations or for SL applications, the deployment manager may centrally push AI models to RAN entities.

\subsection{The data engine}\label{sec5:data_engine}
The data engine serves as a centralized hub for managing data and overseeing the lifecycle of AI models in the network.

Data management entails controlling and configuring distributed data generation from RAN entities hosting actuation or inference for each AI-driven RAN function, as illustrated in Figure~\ref{fig:2}. A data handler stores samples of experience generated from the local environment of different RAN entities in a repository, either internal or external to the data engine, and collaborates with the learning engine to provide data for training, validation, and testing. Collectively, samples of local experience deliver the diversity needed to overcome the system's partial observability and generalize AI solutions. Centralizing control and configuration of field data generation is essential to scalability, diversity and, for RL algorithms, proper exploration (cf. Sec.~\ref{subsec:safe_exploration} for more details). 

The LCM functionality assures that AI models meet specified performance, reliability, security, ethical criteria, standards, etc. This process involves a series of quality checks and testing procedures designed to evaluate the model's behavior and performance in various conditions. The LCM function centrally manages drift detection and model performance monitoring, while the computational demand of these operations can be more efficiently distributed to RAN entities hosting the actor or inference functions. This approach avoids continuous streaming of data samples from local RAN entities to the RLE to detect discrepancies between the runtime local data distribution and the overall distribution of the training data. Communication with LCM functionality can be limited to event-based feedback and on-demand data provisioning. Upon receiving and processing feedback related to drift detection or model performance, the LCM functionality may initiate a model update process involving data management and the learning engine. Additionally, the data engine handles storage and version control of algorithmic objects, such as trained models received from the learning engine, data structures (e.g., learning metadata created during training, testing, and validation processes), and unit tests. For certain types of AI algorithms, such as supervised learning (SL) or offline RL, the data engine may also host functionalities typically associated with a feature store~\cite{FeatureStore:2022}. This topic is further discussed in Sec.~\ref{subsec:beyond_RL_algorithms}.

\subsection{Actor and inference}
Actor and inference functions are responsible for executing AI models, generating and collecting training data, and supporting LCM functionalities such as local performance monitoring, local data drift detection, and fallback operations. Specific AI algorithms necessitate tailored functionalities. For example, a RL actor handles action selection, actuation, and reward computation, while a SL inference function provides automated data labeling. RAN entities hosting local actor or inference functions are the ideal sources for generating training data, provided they are supervised and coordinated by the RLE. They can efficiently access and process algorithm-specific information forming training samples, including the input state, selected action, and reward for RL or data labels for SL, using functionalities typically associated with a feature store \cite{FeatureStore:2022}. Reconstructing training samples elsewhere (e.g., at the RLE) would be inefficient, as it would require continuous streaming of raw data and extensive post-processing.

%% file: components_arXiv_R2/Section6.tex
\section{Practical considerations}\label{sec6:practical_considerations}
We next explore how the architecture outlined in Sec.~\ref{sec:scalable_learning} enables a scalable integration of AI in real-world RAN systems and supports different AI algorithms.

\subsection{Algorithms support}\label{subsec:beyond_RL_algorithms}
Separating learning from actor/inference functions broadens the AI algorithm compatibility of the architecture in Figure~\ref{fig:2}. This separation enables centralized control of data generation across distributed RAN entities, benefiting algorithms like SL and unsupervised learning, which can generate experience outside the training loop for tasks like regression, estimation, classification, or offline RL in control applications. Furthermore, for AI algorithms requiring experience generated while training, like off-policy RL for control tasks, this separation makes the architecture capitalize, rather than suffer from, the lag inherently present in RAN systems between the generation (at the actor) and consumption (at the learner) of data.

This decoupling also enables supporting federated learning (FL) types of algorithms. Borrowing from the Gorila architecture \cite{Nair:2015}, actors would host local learning functions, while data provided to the centralized learner consists of gradients or local versions of the centrally trained model. Comparing FL and RL in terms of performance, data management, security, and hardware and software footprint is beyond the scope of this paper.

\subsection{Scalable data generation and management} \label{subsec:safe_exploration}
As discussed in Sec.~\ref{sec4:impact_training}, collecting data from diverse local environments in real-world RANs, each with unique settings and parameters, poses a significant challenge in training a single model to achieve generalization. This is due to the increased need for data and the risk of catastrophic forgetting arising from limited buffer sizes inherent to such diverse systems. A distributed~data collection strategy is crucial to mitigate data bias and increase sample efficiency, promoting superior learning and data generation scalability. The challenge to overcome is to efficiently handle a potentially massive data influx towards the RLE. 

Network functionalities exhibit varying data throughput across protocol layers. L1-L2 tasks (e.g., scheduling) generate samples in milliseconds per UE and cell. L3 functions (e.g., mobility) operate at tens to hundreds of milliseconds, while network optimization (e.g., antenna tilt) is notably~slower. 

\textbf{\textit{Example:}} In a RAN system with $N=10^3$ cells, link adaptation may produce a system-wide data throughput toward the RLE exceeding $S = 10^6$ samples/sec.

Collecting data too frequently from the same sources, like in L1-L2 tasks, may result in strong correlation and strain data ingestion interfaces. Moreover, managing and storing such data over extended periods is arduous and costly.~In~control applications, such high data throughput far exceeds the requirements for policy updates, hindering optimal policy learning.

To address this challenge, the RLE data engine can centrally orchestrate data generation by distributing it spatially (i.e., across RAN entities) and temporally, prompting RAN entities to produce fewer training samples spread over time. Temporal down-sampling mitigates correlation from the same source, while aggregating experience from distributed RAN entities ensures data diversity and mitigates bias, thus enabling to improve model generalization. 

For RL algorithms, this amounts to slowing down policy evaluation and decoupling policy updates from ongoing evaluations, while centrally controlling actors' exploration enables directed exploration in specific RAN areas. To ensure safe field exploration, actors may operate with a trusted policy or rule-based algorithm when not generating training data, safeguarding system performance.

\subsection{Scalable  model LCM}
Centralizing AI learning services, like training and data management, combined with distributed model drift and performance monitoring is necessary to scalable and automated model LCM operations. However, they alone are insufficient without achieving model generalization, at least over RAN environment, enabling to deploy a single (or a few) model(s) per RAN functionality network-wide. This approach reduces the footprint, complexity, and costs of LCM. 

In contrast, specializing AI models to environments, e.g., a model per cell, leads to overseeing numerous models per RAN functionality. While the RLE allows model specialization, e.g., by filtering training data, possibly using metadata identifying data sources, the LCM operations no longer scale effectively.

\subsection{Mapping to RAN architecture} \label{subsec:mapping_to_RAN}
To map the functionalities in Figure~\ref{fig:2} to a RAN architecture, we distinguish between actor/inference functions and the RLE. 
The actor/inference functions deployment is use-case dependent. For AI algorithms replacing L1-L3 control functions, actors may reside in distributed RAN entities (e.g., gNB-CU for L3 or gNB-DU for L1-L2). Alternatively, centralized inference outside RAN entities can provide insights for legacy L1-L3 functions, like a near real-time RIC for ORAN systems. Actor/inference functions may replace network configuration functionalities in higher RAN layers, such as an r/xApp.

The RLE deployment should be use-case agnostic to support all RAN functions replaced by AI. A cloud-based RLE realization, modeled as a standalone support function, would suit most criteria for supporting AI in both cloud-RAN and integrated-RAN systems. We defer the details to future work. 

\subsection{Efficient resource dimensioning}\label{subsec:resource_dimensioning}
Centralizing learning services for distributed AI-driven network functions enhances the effective dimensioning of computational resources. By centralizing computationally intensive tasks, like training, it enables the aggregation of powerful resources, such as GPUs  or utilization of scalable cloud-based services. This approach establishes a future-proof strategy for AI research. Furthermore, centralization allows for the use of less expensive hardware at the system edge, where action and inference occur. This resource allocation strategy promotes a cost-efficient system, optimizing the use of advanced computational power where most needed and conserving resources at the network's edge.

%% file: components_arXiv_R2/Section7.tex
\section{Design concepts evaluation}\label{sec:Evaluations}

We employ our model generalization principles to develop an RL approach for link adaptation (LA). Initially, we focus on generalizing across the RAN environment for LA, specifically for selecting the MCS index for UE transmissions. Subsequently, we expand our design to encompass generalization across the LA action space, simultaneously controlling the MCS index and rank. Lastly, we address generalization across L2 intents, training an agent to optimize LA parameters to meet various UE-specific criteria such as throughput, spectral efficiency, transmission reliability, or latency.

\subsection{Training process}

We implemented an off-policy RL algorithm following Ape-X architecture~\cite{Horgan:2018}, employing actor-learner separation to scale algorithm training. Actors asynchronously pull model updates from the learner and handles multiple parallel simulations generated by a 5G-compliant event-driven simulator.

During training, we generate 3000 training scenarios, each simulating three seconds (i.e., 3000 TTIs) of a 5G-compliant network yielding in average $\approx$25k training samples, and resulting in $\approx$75 millions samples during the whole training process.
We apply domain randomization to training scenarios to ensure data diversity for achieving model generalization. Each training scenario is produced with a 5G-compliant event-driven system simulator, with simulation parameters randomized in terms network deployment, radio conditions, traffic conditions, etc., according to Table~\ref{table:sim_params}.

Each training scenario simulates a Time-Division Duplexing (TDD) 5G system operating at a 3.5GHz carrier frequency, with PHY layer numerology $\mu=0$ and single-user Multi-Input Multi-Output (SU-MIMO) transmission. A training scenario features of three tri-sectorial radio sites, each randomly configured as either MIMO or massive MIMO (mMIMO), based on the antenna array characteristics defined in Table~\ref{table:sim_params}. The radio site configuration is then randomized in terms of location, cell radius, system bandwidth, and downlink transmit power by sampling parameters from Table~\ref{table:sim_params}. The training scenario is further randomized in terms of cell load, type of traffic, indoor/outdoor UE ratios and UE receiver type. To this end, a UE generation process randomly selects the number of UEs with Full Buffer (FB) traffic and Mobile Broadband (MBB) traffic to be dropped in the scenario according to one of the indoor/outdoor probability values in Table~\ref{table:sim_params}. Each MBB UE generates traffic with a wide range of packet size and arrival time according to a packet arrival process and a packet size distribution modelled based on historical data collected in field campaigns. Each UE configuration is then randomized in terms of number of UE antennas, UE speed, and type of UE receiver. The latter parameter emulates the fact that UEs form different manufactures can experience radio conditions differently due differences in their hardware (e.g., antenna array and chipset) and internal receiver algorithms (e.g., for CSI estimation).

\begin{table}[t]
	\caption{RAN environment simulation parameters.} 
	\centering 
	\begin{tabular}{l l l} 
		\toprule[1pt]\midrule[0.3pt]
		\textbf{Parameter} & \textbf{Value range} & \textbf{Description} \\ [0.5ex]
		\midrule
		Duplexing type & TDD & Fixed\\
		Carrier frequency & 3.5 GHz & Fixed \\
		Deployment type & 3-site 9-sector & \\
		Site type & \{MIMO, mMIMO\} & Randomized\\
		Antenna array & 1x2x2 MIMO (4) & Fixed  \\
		&8x4x2 mMIMO (64) & Fixed \\
		Cell radius & \{166, 300, 600, 900, 1200\} m & Randomized \\
		Bandwidth & \{20, 40, 50, 80, 100\} MHz & Randomized\\
		Number of sub-bands & \{20, 106, 133, 217, 273\} & Randomized \\
		DL TX power & \{20, 40, 50, 80, 100\} W  & Randomized \\
		UE antennas & \{2, 4\} & Randomized \\
		Maximum TX rank & \{2, 4\} & As per UE ant. \\
		Maximum DL TX & 5 & Fixed \\
		UE traffic type & \{FB, MBB\} & Randomized \\
		Number FB UEs  & \{1, 5, 10\} & Randomized \\
		Number MBB UEs & \{0, 10, 25, 50, 100, 200, 300\} & Randomized\\
		Speed UE FB & \{0.67, 10, 15, 30\} m/s & Randomized \\
		Speed UE MBB & \{0.67, 1.5, 3\} m/s & Randomized\\
		UE receiver types & \{type0, type1, type2, type3\} & Randomized \\
		Indoor probability & \{0.2, 0.4, 0.8\} & Randomized\\
		\midrule[0.3pt]
	\end{tabular}\label{table:sim_params}
\end{table}

\subsection{Testing process}

To evaluate our design's generalization, we test it against five benchmark scenarios described in Table~\ref{table:benchmarks}, whose conditions are not directly experienced in the training process. The first benchmark is a homogeneous MIMO scenario with UEs generating full buffer traffic, thereby experiencing stable interference conditions. The second and third benchmark are 3-cell homogeneous mMIMO scenarios, with UE generating full buffer and MBB traffic, respectively. The narrower beam transmissions of mMIMO generates more spotty interference, which becomes even more dynamic in case of MBB traffic. The forth benchmark is a wider 9-cell homogeneous mMIMO scenario with mixed traffic (FB and MBB). Finally, the fifth benchmark is a heterogeneous deployment with MIMO and mMIMO cells, mixed traffic, and heterogeneous UE types. Each benchmark is evaluated with 100 random seeds.

\subsection{Generalization over environment}\label{subsec:results_env}
\begin{figure}[t]
    \centering
    \includegraphics[width=1\columnwidth]{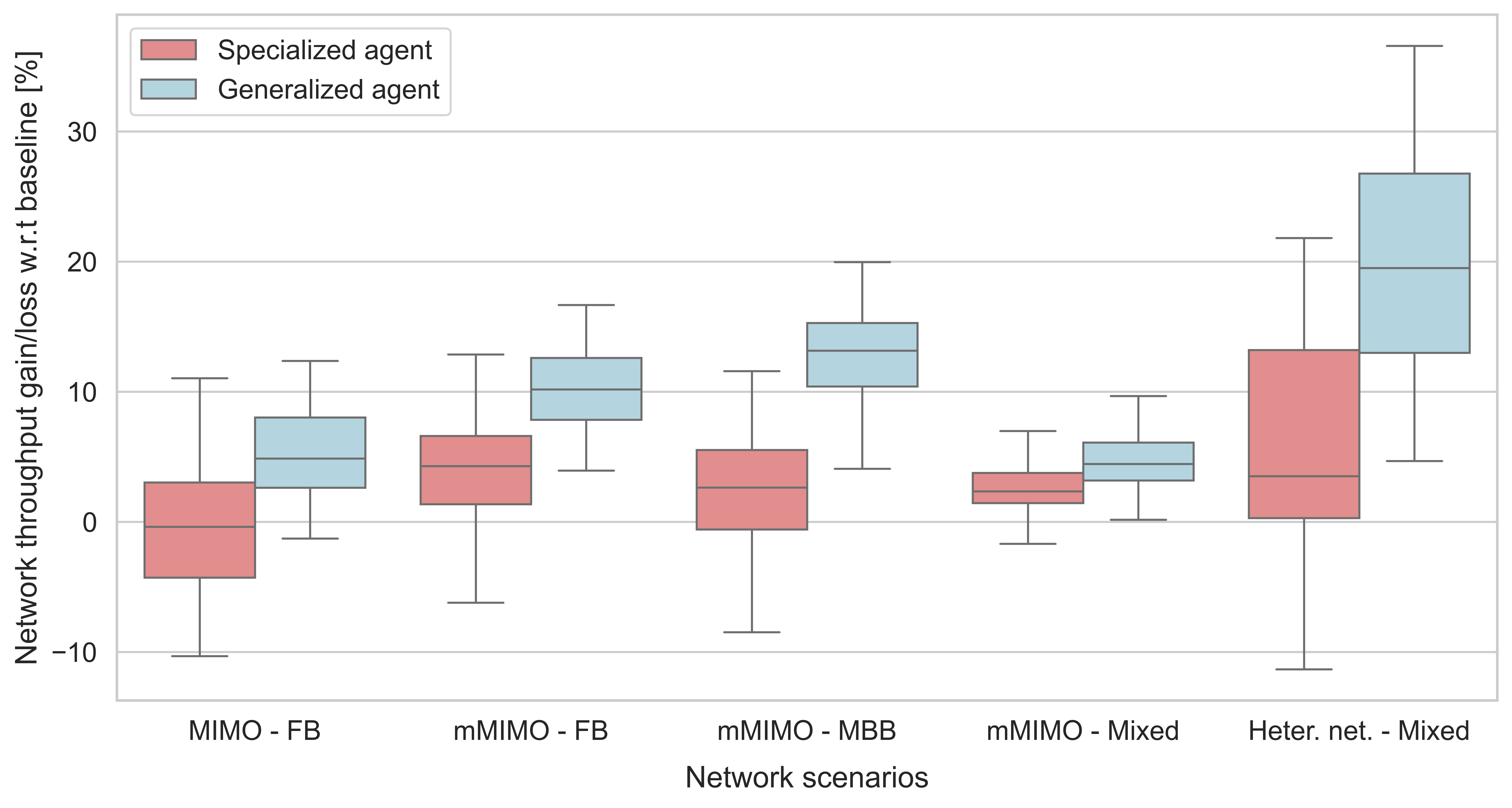}
    \caption{Performance of model generalized over network environment in benchmark scenarios featuring MIMO, mMIMO, and heterogeneous deployments with full buffer (FB), mobile broadband (MBB), or mixed traffic.}
    \label{fig:3}
\end{figure}
We design an LA agent to generalize over the environment considering static and dynamic environment information available within a UE-centric neighborhood, as in Figure~\ref{fig:1}. Static information characterizes the UE-centric neighborhood formed by its serving cell and two relevant interfering cells, as well as information characterizing the UE and its capabilities. Dynamic information comprises radio measurements, channel state information, HARQ feedback, UE buffer state, and more. Our evaluation focuses on SU-MIMO, facilitating a scheduling-agnostic algorithm comparison. 

Figure~\ref{fig:3} compares the generalized agent to a specialized agent trained exclusively for a heterogeneous deployment of MIMO and massive MIMO (mMIMO) using a $10\%$ BLER target outer-loop LA algorithm as a baseline. The generalized agent consistently outperforms the baseline and the specialized agent across all benchmarks, achieving over $20\%$ network throughput gains in challenging scenarios and smaller improvements in cases where the baseline performs well, like full-buffer traffic. Figure~\ref{fig:3} highlights the limitations of training specialized models, resulting in less adaptable designs for diverse RAN environments.

The generalized agent's superior performance can be attributed to an enhanced feature design, e.g., incorporating input features such as the type of serving and neighboring cells. Moreover, its training involved exposure to an extensive range of random environment parameters, encompassing varying user loads, downlink traffics, interference levels, and more, spanning from low to moderate and high values. This extensive training enabled the generalized agent to outperform the specialized agent, even in scenarios closely resembling those on which the latter agent was initially trained, as illustrated in Figure~\ref{fig:3} for mMIMO scenarios.

\subsection{Generalization over control}\label{subsec:results_control}
\begin{figure}[t]
    \centering
    \includegraphics[width=1\columnwidth]{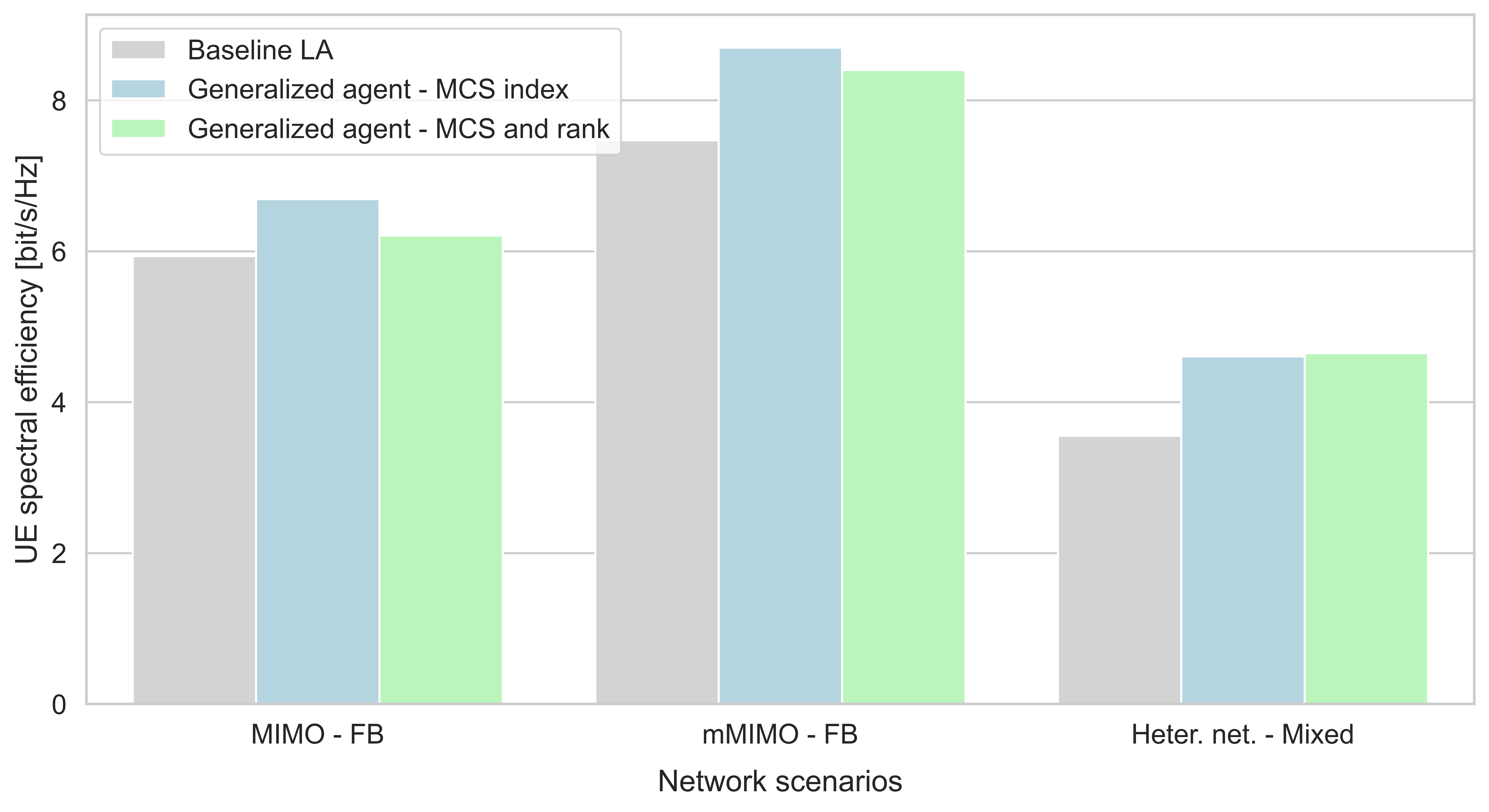}
    \caption{Evaluation of model generalization over LA control actions in an SU-MIMO setup.}
    \label{fig:4}
\end{figure}

Generalizing our design over LA actions to concurrently control MCS index and transmission rank increases the action space, an exponential state-action space growth. We trained a new agent with this extended action space, using the same state and reward design and training methodology. However, we observed no significant performance improvement compared to using the UE-reported rank as part of the CSI (see Figure~\ref{fig:4}). This example demonstrates that generalization over LA actions is not justified for SU-MIMO transmission. However, it could yield substantial gains in multi-user MIMO scenarios. Future research will explore this avenue further.

\subsection{Generalization over intents}\label{subsec:results_intents}
We devised an LA algorithm to generalize over MAC layer KPIs employing multi-objective RL with envelope Q-learning \cite{YSN:19}. Our design features a 2-dimensional reward $r$ comprising information bits and total resource usage for packet delivery. A preference vector $\omega$ assigns weights to reward components based on user intents. Through training with various preference values, we obtained a single policy that optimizes the Pareto frontier for all reward components. Figure~\ref{fig:5} shows the design's performance in terms of user throughput, spectral efficiency, and BLER across preference values $\omega$. At runtime, choosing $\omega$ based on user intents allows executing this policy to optimize specific behaviors or KPIs (e.g., 5QI QoS parameters) on a per-UE basis. Choosing $\omega\approx 0$ enhances reliability and latency (e.g., for URLLC traffic), while $\omega\approx 1$, maximizes spectral efficiency but incurs higher latency. Optimizing UE throughput requires balancing preferences among reward components.

%% file: components_arXiv_R2/Section8.tex
\section{Conclusions}\label{sec:conclusions}
We present AI model generalization as fundamental requirement for a scalable integration of AI in RAN systems, and identify three primary design domains pertinent to model generalization in RAN systems: \emph{RAN environment} to enable ubiquitous deployment of a single model with consistent performance; \emph{RAN intents} to handle and optimize different key performance indicators (KPIs) with a single model at runtime; and \emph{RAN control} to reduce the number of co-existing AI-driven control loops acting on interdependent RAN functionalities. We discuss the challenges that achieving model generalization poses to the scalability of training and data generation, and we propose an architecture that addresses these challenges effectively and suits radio communication systems well. The architecture features centralization of training and data management functionalities, combined with distributed data generation across RAN entities hosting AI functionalities. These concepts are demonstrated through a generalized link adaptation algorithm.

\begin{figure}[t]
    \centering
    \includegraphics[width=1\columnwidth]{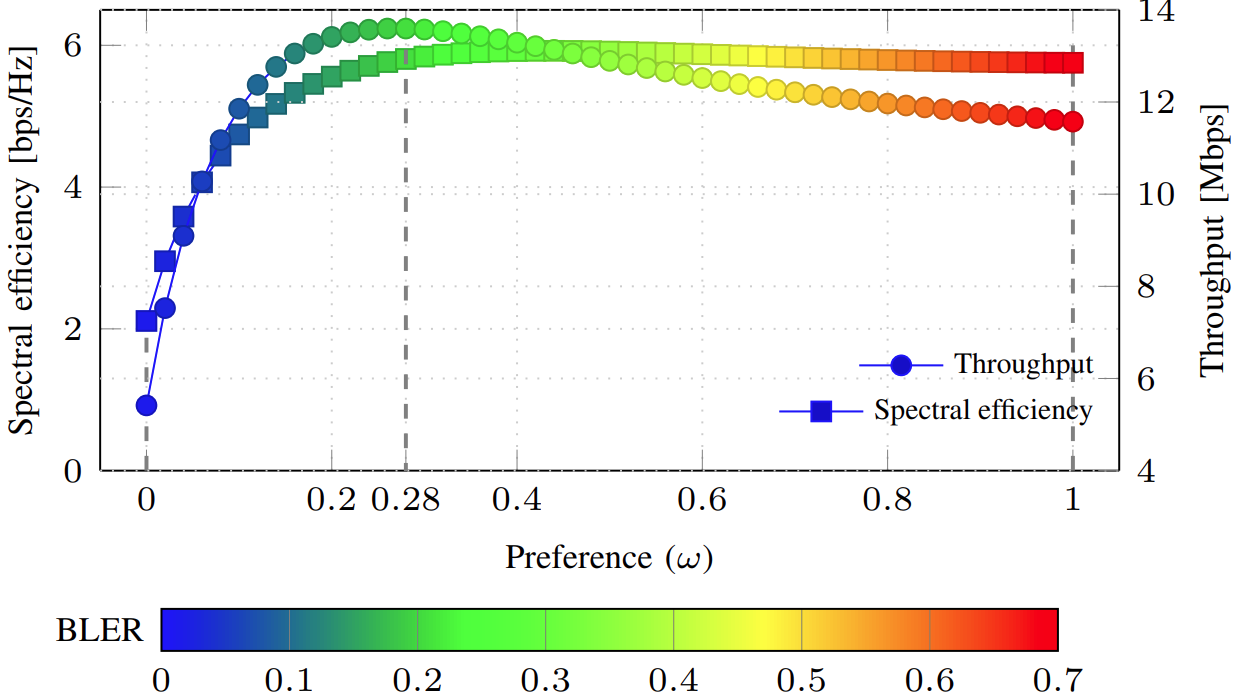}
    \caption{Evaluation of model generalization over MAC intents to flexibly adapt LA parameters to meet specific intents.}
    \label{fig:5}
\end{figure}

\begin{table*}[t!]
	\caption{Benchmark testing scenarios} 
	\centering 
	\begin{tabular}{c l l l l} 
		\toprule[1pt]\midrule[0.3pt]
		\textbf{Name} & \textbf{Type} & \textbf{Description} & \textbf{Traffic} & \textbf{User types}  \\ [0.5ex]
		\midrule
		MIMO-FB & Stable interference & 1-site 3-cell MIMO, $80\%$ indoor  UEs& Full buffer & Homogeneous (type0)\\
		mMIMO-FB& Stable interference & 1-site 3-cell mMIMO, $80\%$ indoor UEs & Full buffer & Homogeneous (type0)\\
		mMIMO-MBB& Dynamic interference & 1-site 3-cell mMIMO, $80\%$ indoor UEs & Mobile broadband & Homogeneous (type0)\\
		mMIMO-Mixed& Mixed traffic & 3-site 9-cell mMIMO, $80\%$ indoor UEs & Mixed MBB and FB & Homogeneous (type0)\\
		Het.Net.-Mixed& Randomized scenarios & 3-site 9-cell heterogeneous networks based on Table~\ref{table:sim_params} & Mixed MBB and FB  & Heterogeneous\\
		\midrule[0.3pt]
	\end{tabular}\label{table:benchmarks}
\end{table*}

%% file: components_arXiv_R2/Biographies.tex
\section{Biographies}
\vspace{-33pt}
\begin{IEEEbiographynophoto}{Pablo Soldati}
is a Standardization and Concepts Researcher at Ericsson for AI integration in radio networks. His research interests include AI/ML, optimization theory and wireless networks. He holds a Ph.D. in telecommunications from KTH Royal Institute of Technology, Sweden. 
\end{IEEEbiographynophoto}

\vspace{-33pt}
\begin{IEEEbiographynophoto}{Euhanna Ghadimi}
is an Expert at Ericsson in AI algorithms for radio access networks. Alongside AI/ML, he also has research interests in optimization theory and wireless networks. He holds a Ph.D. in telecommunications from KTH Royal Institute of Technology, Sweden. 
\end{IEEEbiographynophoto}

\vspace{-33pt}
\begin{IEEEbiographynophoto}{Burak Demirel}
is a Senior Researcher at Ericsson. His work focuses on AI/ML, control theory, and cyber-physical systems. He received his Ph.D. in automatic control from KTH Royal Institute of Technology, Sweden. 
\end{IEEEbiographynophoto}

\vspace{-33pt}
\begin{IEEEbiographynophoto}{Yu Wang}
is a Concept Developer at Ericsson for development of AI/ML-based radio network automation solutions. His research interests include RRM, network management and telecom data analytics. Wang holds a Ph.D. in communication engineering from Aalborg University, Denmark.
\end{IEEEbiographynophoto}

\vspace{-33pt}
\begin{IEEEbiographynophoto}{Raimundas Gaigalas}
is a Concept Developer for Cloud RAN at Ericsson for AI/ML functions in the RAN products. His research interests include development and systemization of 4G/5G RRM commercial features. Gaigalas holds a Ph.D. in mathematical statistics from Uppsala University, Sweden. 
\end{IEEEbiographynophoto}

\vspace{-33pt}
\begin{IEEEbiographynophoto}{Mathias Sintorn}
is an Expert at Ericsson in traffic handling and service performance within Business Area Networks. He defines the long-term evolution of the RAN architecture, specifically in RAN automation. Sintorn holds an M.Sc. in engineering physics from Uppsala University, Sweden.
\end{IEEEbiographynophoto}

\vfill